\documentclass[11pt]{article}

\usepackage[preprint]{acl}

\usepackage{times}
\usepackage{latexsym}

\usepackage[T1]{fontenc}

\usepackage[utf8]{inputenc}

\usepackage{microtype}

\usepackage{inconsolata}

\usepackage{graphicx}
\usepackage{alltt}
\usepackage{multirow}

\usepackage{subcaption}
\usepackage{amsmath}
\usepackage{placeins}
\usepackage{float}
\usepackage{booktabs}
\usepackage{makecell}

\usepackage{listings}
\usepackage{xcolor}
\usepackage[export]{adjustbox}

\usepackage{xcolor}
\definecolor{mygray}{rgb}{0.95,0.95,0.95}
\lstdefinestyle{regexstyle}{
  basicstyle=\ttfamily\small\color{purple},
  breaklines=true,
  backgroundcolor=\color{gray!10},
  frame=single
}

\definecolor{bg}{rgb}{0.97,0.97,0.97}
\definecolor{string}{rgb}{0.58,0,0.82}
\definecolor{comment}{rgb}{0,0.5,0.2}
\definecolor{keyword}{rgb}{0.13,0.13,0.65}

\lstdefinestyle{pretty}{
  backgroundcolor=\color{bg},
  frame=single,
  rulecolor=\color{gray},
  numbers=left,
  numberstyle=\tiny\color{gray},
  basicstyle=\ttfamily\small,
  keywordstyle=\color{keyword}\bfseries,
  stringstyle=\color{string},
  commentstyle=\color{comment}\itshape,
  breaklines=true,
  tabsize=2,
  captionpos=b,
  aboveskip=1em,
  belowskip=1em,
}

\lstdefinestyle{prettyyaml}{
  basicstyle=\ttfamily\footnotesize,
  backgroundcolor=\color{bg},
  frame=single,
  rulecolor=\color{gray},
  numbers=none,
  breaklines=true,
  tabsize=2,
  captionpos=b,
  aboveskip=0.7em,
  belowskip=0.7em,
  xleftmargin=0.08\textwidth, }

\usepackage[framemethod=TikZ]{mdframed}
\usepackage{xcolor}

\newmdenv[
  backgroundcolor=gray!7,
  linecolor=gray!50,
  roundcorner=4pt,
  skipabove=0.7em,
  skipbelow=0.7em,
  innertopmargin=0.7em,
  innerbottommargin=0.7em,
  innerleftmargin=1em,
  innerrightmargin=1em,
  linewidth=0.7pt
]{edubox}

\usepackage[textsize=scriptsize, colorinlistoftodos]{todonotes}
\usepackage{changes}

\newif\ifshowdg \showdgfalse   \newif\ifshowvhan \showvhanfalse   \newif\ifshowvhaag \showvhaagfalse  \newif\ifshowlh \showlhtrue   \newif\ifshowas \showastrue   \newif\ifshowjs \showjstrue   
\newcommand{\dg}[1]{  \ifshowdg
    \todo[linecolor=magenta, backgroundcolor=magenta!10]{\textbf{DG:} #1}  \fi
}

\newcommand{\vhaag}[1]{  \ifshowvhaag
    \todo[linecolor=yellow, backgroundcolor=yellow!10]{\textbf{VHaag:} #1}  \fi
}

\usepackage{listings}
\usepackage{xcolor}
\usepackage{enumitem}

\lstdefinelanguage{yaml}{
    keywords={true,false,null},
    sensitive=true,
    comment=[l]{\#},
    morestring=[b]",
}

\usepackage[acronym]{glossaries}
\usepackage{float}
\glsdisablehyper
\makeglossaries

\newacronym{llm}{LLM}{large language model}
\newacronym{ai}{AI}{artificial intelligence}
\newacronym{pii}{PII}{personally identifiable information}
\newacronym{cc}{CC}{Common Crawl}
\newacronym{lr}{LR}{learning rate}
\newacronym{arc}{ARC}{AI2 Reasoning Challenge}
\newacronym{mmlu}{MMLU}{Massive Multitask Language Understanding}
\newacronym{gsm8k}{GSM8K}{Grade School Math 8K}
\newacronym{ifeval}{IFEval}{Instruction-Following Eval}
\newacronym{m}{M}{million}
\newacronym{b}{B}{billion}
\newacronym{t}{T}{trillion}
\newacronym{fwe-de}{FWe\text{-}de}{FineWebEdu German}
\newacronym{gelu}{GELU}{Gaussian Error Linear Units activation}
\newacronym{mlp}{MLP}{Multilayer Perceptron}
\newacronym{LayerNorm}{LayerNorm}{Layer normalization}
\newacronym{NFKC}{NFKC}{Normalization Form Compatibility Composed}
\newacronym{PLRS}{PLRS}{power-law learning rate schedule}
\newacronym{FSDP}{FSDP}{Fully Sharded Data Parallel}
\newacronym{SFT}{SFT}{Supervised Fine-Tuning}

\usepackage{hyperref}
\usepackage{xurl}

\title{From Data to Device: ELMOD \\ An Efficient German-First 2.7B Language Model for Mobile Inference}

\author{
Darina Gold\thanks{These authors contributed equally to this work.},
Alexander Schwirjow\footnotemark[1],
Viktor Haag\footnotemark[1],  \\
\bf{Viktor Hangya\footnotemark[1]},
Joel Schlotthauer\footnotemark[1],
Fabian Küch \and 
Luzian Hahn \\
IIS Fraunhofer \\
\texttt{firstname.secondname@iis.fraunhofer.de}
}

\begin{document}
\maketitle
\begin{abstract}
We present ELMOD — Efficient Language Model for On-Device Deployment — a compact (2.7B) German language model designed for efficient inference on resource-constrained hardware. 
ELMOD was trained on a limited computational budget (55k H100 GPU hours)
using exclusively publicly available data.
We developed a suite of German-specific data pre-processing, which differ from English-oriented counterparts in their handling of morphological variation, compounding, and orthographic conventions. Furthermore, we introduced a quality filtering and rephrasing step, which increased the instructional quality of the data, improved performance during the annealing phase, and reduced overall compute requirements.
Thanks to our architectural model and data choices, 
including prefiltering, our educational-quality filtering and rephrasal to raise the educational-quality, 
ELMOD is the strongest performer in its size class (<3B), matching the performance of 7B-parameter models in German.
\end{abstract}

\section{Introduction}

Scaling language models with more data, parameters, and compute has enabled impressive gains in language modeling capability. \cite{yang2025qwen3technicalreport,kimiteam2026kimik2openagentic,nvidia2026nemotron3ultraopen}
Yet, these gains are accompanied by growing reliance on remote servers, persistent connectivity, and energy-intensive infrastructure, limiting reliability, accessibility, and user privacy in real-world deployment.
This paper presents a comprehensive methodology for building an efficient, German and English language model on a limited compute budget (55k H100 GPU hours). The process covers data collection, prefiltering, text quality filtering, and training, with ablation studies at each stage shaping the final model design.

In contrast to this, major LLM providers treat non-English languages, such as German as a by-product and do not prioritize local deployment, while following European regulations. The EU-AI Act enforces several criteria of transparency, which are not covered by other models below 3B-Parameters at the current point in time.

We detail the data preparation process and the full training pipeline, and introduce ELMOD-2.7B, an efficient and effective German language model that carefully balances performance and resource efficiency. 
With 2.7 \gls{b} parameters trained on 4 \gls{t} tokens, 
our model is sufficiently compact to enable deployment on modern mobile and edge devices.
ELMOD-2.7B achieves the best performance within its size class and matches the performance of significantly larger 7B-parameter models on German language benchmarks (the German versions of \gls{arc}
, HellaSwag, and \gls{mmlu}). 
To the best of our knowledge, we are the first to outline the model creation process from data collection to on-device model deployment for a German model.\footnote{\url{https://huggingface.co/collections/fraunhofer-iis/elmod-27b}}

In the description of our model development, we focus on three principal aspects (see Figure~\ref{fig:overview} for an overview) : 
\vspace{-10pt}
\paragraph{Data Preparation:} 
Design and implementation of the pre-processing pipeline (see Section~\ref{sec:pre-training_data}), quality classification and filtering, and reformulation of low-quality texts into high-quality equivalents.
\vspace{-10pt}
\paragraph{Pre-training} Description of our pre-training (see Section~\ref{sec:pre-training}), our primary training focus, including key design choices and ablations (e.g., tokenizer and learning-rate annealing).
\vspace{-10pt}
\paragraph{Post-training}
We also train an instruction-tuned model with performance comparable to similar models,
although we do not perform extensive post-training optimizations,
and deploy ELMOD on-device as a demonstration (see Section~\ref{sec:post_training}).

\section{Related Work}
In this section, we review the landscape of German language models (see Section~\ref{sec:rl_german_models}) and describe data selection for pre-training (see Section~\ref{sec:rl_data}).

\subsection{German and German-Capable Language Models}
\label{sec:rl_german_models}
While several German-capable \glspl{llm} already exist \cite{pluester2023b, ostendorff2023efficient, delobelle2024buble, shliazhko2024mgpt, mistral7b, ali2024teuken, pfister-etal-2025-llammlein, burns2025aleph}
, these efforts 
rarely provide a comprehensive, step-by-step guide for building models from scratch in low-resource settings, particularly for non-English.

Several German-focused \glspl{llm} have been introduced, including models fine-tuned on German data (e.g., LeoLM \cite{pluester2023b}, BübleLM \cite{delobelle2024buble}, bloom-clp \cite{ostendorff2023efficient}) and others trained from scratch (e.g., mGPT \cite{shliazhko2024mgpt}, Mistral-7B \cite{mistral7b}, Teuken-7B \cite{ali2024teuken}, Llämmlein \cite{pfister-etal-2025-llammlein}, Aleph-Alpha-Web \cite{burns2025aleph}). 
However, while mGPT and Mistral-7B lack detail about their German training data, Teuken-7B, Llämmlein, and Aleph-Alpha-Web provide greater transparency, enabling a direct comparison with our approach.

\subsection{Data choice for pre-training}
\label{sec:rl_data}

\paragraph{Data filtering for model training}
\label{sec:rl_data_filtering}
Recent work has shown that improving data quality can substantially enhance training efficiency, enabling models to achieve higher performance with considerably less data \cite{marion2023less,su2024nemotron,burns2025aleph}.
According to \citet{burns2025aleph}, improving data quality has shifted from whitelisting high-quality sources to curating datasets via large-scale filtering and quality scoring of web crawls, retaining coherent, factual, educational content while removing low-quality or redundant data.

\paragraph{Data sources}
\citet{li2025rethinking} show continual pre-training on bilingual data and code multilingual representation learning, while \citet{zhang-etal-2024-plug} demonstrate that leveraging a high-resource pivot language such as English improves the performance of other languages within the model.
Furthermore, \citet{petty2025doescodepretrainingaffect} provide evidence that incorporating code in the training data of a model improves performance on compositional tasks involving structured output, and mathematics.

\section{Data Preparation}
\label{sec:pre-training_data}
\begin{figure*}
    \centering
    \includegraphics[width=0.75\linewidth]{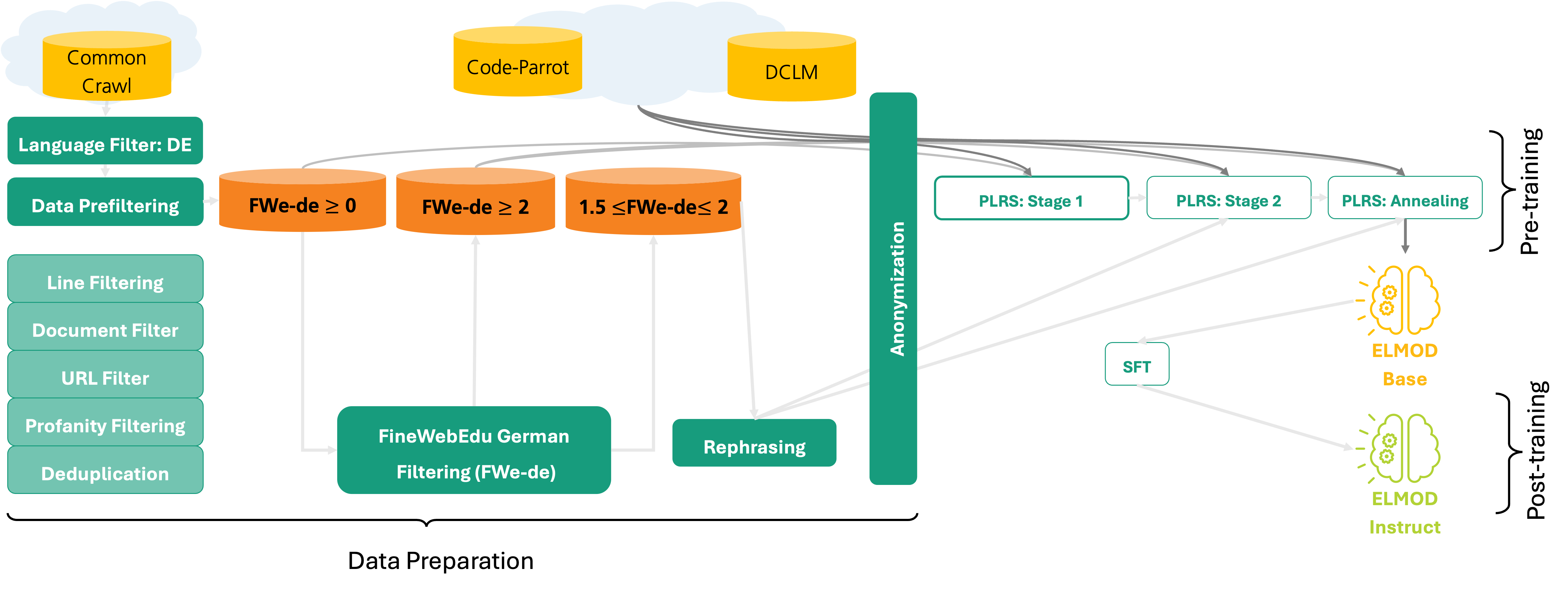}
    \caption{Overview over model creation process}
    \label{fig:overview}
\end{figure*}

Figure~\ref{fig:overview} presents an overview of our model creation process, beginning with the data preparation pipeline. 
The first step to be able to do any data preparation is selecting and downloading the sources, which in our case is web data (see Section~\ref{sec:data_sources}).
Motivated by the previously mentioned developments in data quality and pre-training dataset design, we apply data curation filtering (see Section~\ref{sec:data_filtering}), focused on ensuring dataset suitability rather than quality, in the next step.
Although the general outline to data filtering is established, we are the first ones to fully outline and build such a pipeline from scratch for German.
Furthermore, the steps and their order are designed with another overarching goal of this paper: to save computing resources.
In the third pre-processing step, we build an educational text quality filter classifier with the goal to effectively use our compute especially in the annealing phase (see Section~\ref{sec:lrs} for the annealing phase). 
We discuss quality bucketing in Section~\ref{sec:quality_bucketing}.
Lastly, we augment high educational quality data through synthesis on the German portion of our pre-training dataset (see Section~\ref{sec:synthetic_data_generation}).
Few works report anonymizing model training data \cite{lothritz-etal-2023-evaluating}, yet to our knowledge, we are the first to explicitly describe it in data preparation and in a model training pipeline overall (see Section~\ref{sec:data_anonymization}).

\subsection{Data sources}
\label{sec:data_sources}

Based on previous work described in Section~\ref{sec:rl_data}, our dataset comprises 45\% German data, 45\% English data, and 10\% code to build a German-capable model.

\paragraph{CodeParrot}
For the code portion, we use the cleaned version of CodeParrot\footnote{\url{https://huggingface.co/datasets/codeparrot/github-code-clean}}, a large-scale dataset of open source GitHub source files spanning 30 programming languages. 
Its breadth of languages and coding styles provides a diverse set of structured text, supporting the model in developing general language understanding, abstraction, and logical reasoning capabilities.

\paragraph{\gls{cc}}
\gls{cc}\footnote{\url{https://commoncrawl.org/}} is one of the largest publicly available web-scale text datasets, containing hundreds of terabytes of raw web data across many languages. 
Additionally, \gls{cc} offers high diversity across domains—ranging from news, blogs, and scientific articles to forums, reviews, and Wikipedia—helping reduce domain bias and improve model robustness.
Finally, it 
has been widely adopted in prior multilingual and large-scale language model work (e.g., mT5 \cite{xue-etal-2021-mt5}, XLM-R \cite{conneau-etal-2020-unsupervised}, CCNet \cite{wenzek-etal-2020-ccnet}), making it a practical and well-understood resource.
We use DCLM\footnote{a random portion taken from DataComp-LM \cite{li2024datacomp}} for the English data and raw \gls{cc} dumps for the German data, covering Common Crawl snapshots up to the March 2025 cutoff (4.83T words in total), and filter it as described in the following section.

\subsection{Data Filtering}
\label{sec:data_filtering}
For our filtering, we apply rule- and gazetteer-based methods to remove non-target languages, low-information text, and profane content. 
These methods are computationally inexpensive compared with learning-based approaches, including both classic classifiers and neural models, and were therefore applied in the early stages.
First, we process 107 WARC dumps, extract the text using ChatNoir Resiliparse \cite{bevendorff:2018} and then go through the filtering procedure as described in this section.
In total 4.83\gls{t} words are reduced to 0.94\gls{t} words (for yield per step, see Table~\ref{tab:document_yield_data_filtering} in the Appendix).

We apply the following filtering steps sequentially, ordering them to minimize computational cost by eliminating as much data as possible in early stages so that texts discarded later are not repeatedly processed:

\paragraph{Language Filter}
To collect German data, we filter for German content during download using the language label provided by Resiliparse\footnote{We use the language code rather than the language score.}.

\paragraph{Line Filter}
We use line- and document filters similar to the C4 approach \cite{dodge2021documenting}\textemdash adjusted to German, meaning that we e.g. translate the key words that are used for filtering and adjust the upper and lower word ratio. Similar C4-inspired heuristic filtering is also employed in several recent web-scale pre-training corpora, including FineWeb \cite{penedo2024finewebdatasetsdecantingweb} and RedPajama-V2 \cite{weber2024redpajamaopendatasettraining}. 
These filters remove lines that contain little or no meaningful natural-language content, thereby reducing noise while retaining informative text. 
Our line filter removes noisy web text by discarding lines that are mostly numbers or uppercase, lines that are repetitive or highly duplicated, and lines that contain certain keywords too often. 
It also applies sequence-based removal for repeated short sequences. The filter is displayed in the Appendix (see Listing~\ref{lst:line_filter}).

\paragraph{Document Filter}
Our document filter is inspired by the Gopher model and its MassiveText dataset \cite{Rae2021ScalingLM}.
Similar Gopher-inspired document filtering heuristics are employed in several recent web-scale pre-training corpora, Dolma \cite{soldaini2024dolmaopencorpustrillion} and RedPajama-V2 \cite{weber2024redpajamaopendatasettraining}, to remove documents with little meaningful natural-language content while retaining high-quality text.
The document filter removes documents based on specific criteria: 
it enforces a minimum number of words, constrains average word length, requires a minimum fraction of valid words and stopwords, and limits punctuation-heavy content such as bullet points or ellipses. 
We also apply a stop word-based filter (by adjusting the stop word list and using the German one from spaCy\footnote{\url{https://github.com/explosion/spaCy/blob/master/spacy/lang/de/stop_words.py}}) to ensure that the text follows natural language patterns. 
Besides the stop word list adjustment, we also adjust the metrics with word lengths to German.
The filter is displayed in the Appendix (see Listing~\ref{lst:doc_filter}).

\paragraph{URL Filter}
We use the UT1 list\footnote{\url{https://dsi.ut-capitole.fr/blacklists/index_en.php}} as a URL blacklist to exclude URLs associated with known undesirable or harmful content categories (e.g., adult, ads, malware, phishing, fake news, etc.).
URL blocklists are commonly used during web-scale corpus construction to filter low-quality or potentially harmful sources before text extraction, thereby improving the overall quality of the resulting corpus \cite{dodge2021documenting,penedo2024finewebdatasetsdecantingweb}

\paragraph{Profanity Filter}
We filter out texts exceeding a defined profanity score (ratio of profane words) to help reduce problematic content and improve training quality and model behavior using the German and English version of the List-of-Dirty-Naughty-Obscene-and-Otherwise-Bad-Words (LDNOOBW)\footnote{\url{https://github.com/LDNOOBW/List-of-Dirty-Naughty-Obscene-and-Otherwise-Bad-Words}}.
Similar filtering strategies have been considered in large-scale pre-training corpora; for example, the C4 data processing pipeline includes an optional bad-words filter based on LDNOOBW lists \cite{dodge2021documenting}. 
Unlike this document-level filtering approach, we apply a ratio-based threshold to avoid removing documents containing only isolated profane terms.
The threshold is set to 0.005.
\paragraph{Deduplication}
Consistent with our overall principle of reducing compute and cost while preserving efficiency, we use deduplication to eliminate redundant documents from the training corpus, while also minimizing the risk of memorization \cite{lee2022deduplicating}.
We use Bloom Filter \cite{bloom1970space} and MinHash deduplication \cite{broder1997resemblance} on per-dump basis. 
We apply a deduplication filter after Bloom filtering, following the approach of \citet{penedo2024finewebdatasetsdecantingweb} and using 128 permutations split into 9 buckets with 13 rows each, a similarity threshold of 0.8. 
As in FineWeb \cite{penedo2024finewebdatasetsdecantingweb}, this configuration keeps computational complexity low while providing effective deduplication.
(For a detailed example of the process see Table~\ref{tab:dedup_example}.) 

\subsection{Quality bucketing data}
\label{sec:quality_bucketing}
Inspired by \citet{su2024nemotron, lozhkovfineweb} and \citet{burns2025aleph}, we employ educational classification to filter low-quality data, a strategy that seemed particularly promising under our compute constraints, as it may allow models to learn more efficiently from higher-quality examples. 
In line with their methodology, we first use an LLM-as-a-judge to annotate a subset of our pre-training corpus, and subsequently trained a regression model to assess data quality across the entire pre-training dataset. 
In the following experiments, we confirm that prioritizing educationally higher-quality data indeed leads to improved training efficiency and model performance in our scenario.

\paragraph{LLM-as-a-Judge Data Scoring}
Consistent with \citet{su2024nemotron} and \citet{burns2025aleph}, we employ 
\textit{Mistral-Small-24B-Instruct-2501} to assess the educational quality of text on a 0–5 scale.
The prompts used for this evaluation are provided in Appendix~\ref{lst:app_promps_edu}. 
We apply this procedure to 1.4 \gls{m} documents sampled from a Common Crawl dump.

\begin{figure}
    \centering
    \includegraphics[width=1\linewidth]{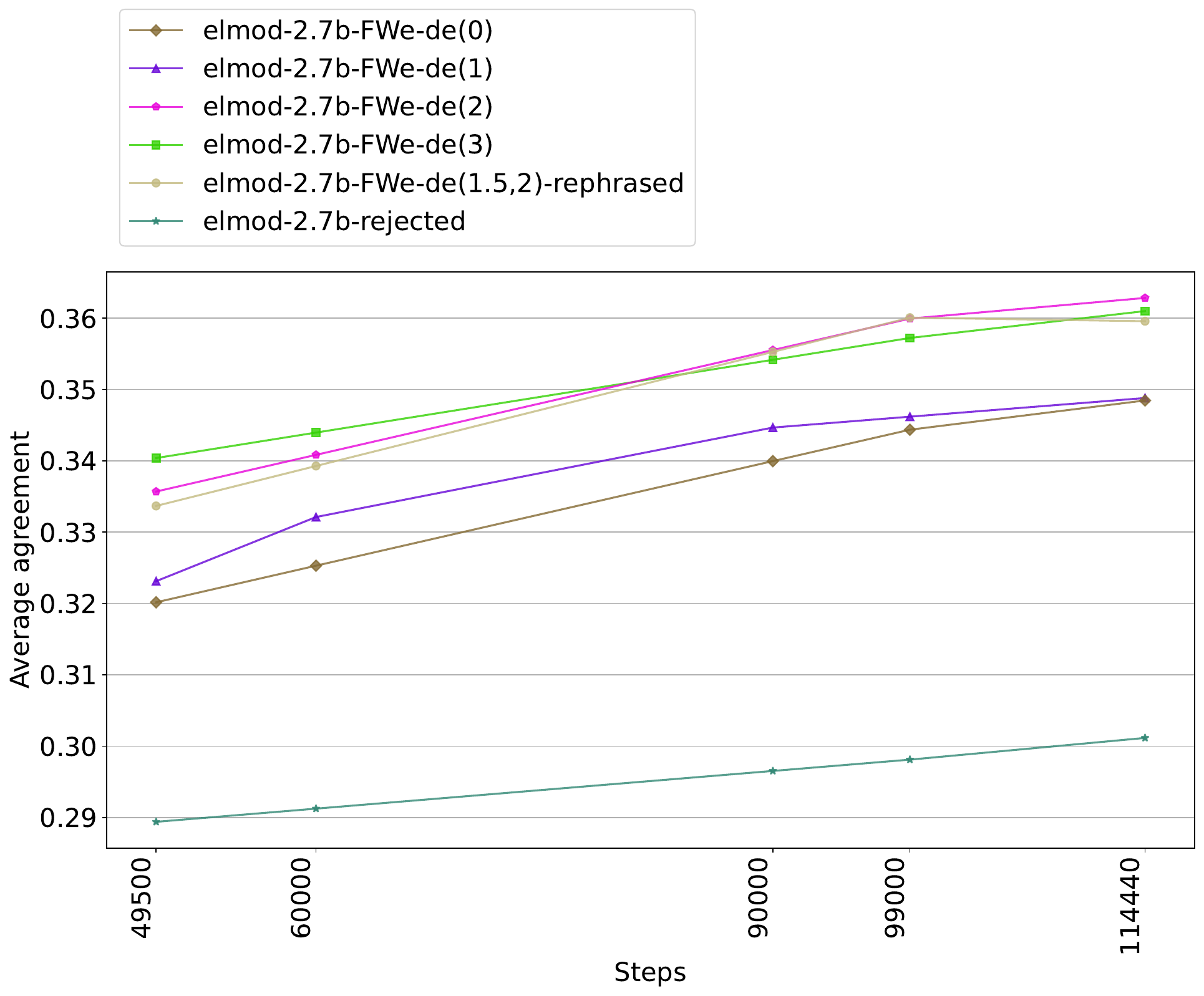}
    \caption{Average model performance on data of varying quality on German tasks:  rejected data (\textsc{elmod-2.7b-rejected}) performs worst, while pre-processed data (\textsc{elmod-2.7b-$\mathrm{\gls{fwe-de}(0)}$}) yields better results. 
    No improvement from \textsc{elmod-2.7b-$\mathrm{\gls{fwe-de}(1)}$} over \textsc{elmod-2.7b-$\mathrm{\gls{fwe-de}(0)}$}.
    Performance improves with $\mathrm{\gls{fwe-de}(2)}$, but not beyond $\mathrm{\gls{fwe-de}(3)}$. 
    Rephrased data (\textsc{elmod-2.7b-$\mathrm{\gls{fwe-de}(1.5,2)}$-rephrased}) performs on-par with $\mathrm{\gls{fwe-de}(2)}$ and $\mathrm{\gls{fwe-de}(3)}$.
    }
    \label{fig:rephrased_avg_over_de_tasks}
\end{figure}

\paragraph{Regression-Based Data Quality Estimation}
Similar to \citet{su2024nemotron} and \citet{lozhkovfineweb}, we use the data annotated by the LLM-as-a-judge to train a linear regression classifier based on \textit{snowflake-arctic-embed-m} text embeddings to estimate the educational quality of texts.
Being based on FineWeb-Edu by \citet{lozhkovfineweb}, we call the data quality scores resulting from our adapted classifier \gls{fwe-de}. 
Throughout this paper, we use the notation $\mathrm{\gls{fwe-de}(X)}$ to denote $\mathrm{\gls{fwe-de} \ge X}$, and $\mathrm{\gls{fwe-de}(X,Y)}$ to denote $Y \ge \mathrm{\gls{fwe-de} \ge X}$. 
The largest portion of the corpus (53 \%) was assigned $\mathrm{\gls{fwe-de}(0,1)}$, indicating non-educational content.\footnote{Table~\ref{tab:llm_filter_yield} presents the resulting distribution of data quality scores assigned by our classifier (\gls{fwe-de})}
In order to show the impact of training data quality on average German benchmark performance, we train six models on 20 billion tokens respectively: \textsc{elmod-2.7b-$\mathrm{\gls{fwe-de}(0)}$} (based on data that came out of the prefiltering pipeline, without further filtering through \gls{fwe-de}), \textsc{elmod-2.7b-$\mathrm{\gls{fwe-de}(1)}$}, \textsc{elmod-2.7b-$\mathrm{\gls{fwe-de}(2)}$},  \textsc{elmod-2.7b-$\mathrm{\gls{fwe-de}(3)}$}, \textsc{elmod-2.7b-rejected} (based on data that is filtered out after the \textit{document filtering} step) and
\textsc{elmod-2.7b-$\mathrm{\gls{fwe-de}(1.5,2)}$-rephrased} (based on rephrased data which will be introduced and discussed in detail in Section~\ref{sec:synthetic_data}).
Figure~\ref{fig:rephrased_avg_over_de_tasks} shows these results.

The performance gain of \textsc{elmod-2.7b-rejected} shows that even subpar data enables learning progress.
Furthermore, it shows that data filtered through straightforward filtering methods, such as regular-expression and gazetteer-based steps leads to a considerable model performance gain (\textsc{elmod-2.7b-$\mathrm{\gls{fwe-de}(0)}$}).

While \textsc{elmod-2.7b-$\mathrm{\gls{fwe-de}(1)}$} is on par with \textsc{elmod-2.7b-$\mathrm{\gls{fwe-de}(0)}$}, models trained on \textsc{elmod-2.7b-$\mathrm{\gls{fwe-de}(2)}$} perform measurable better than these. 
However, there is no greater gain from \textsc{elmod-2.7b-$\mathrm{\gls{fwe-de}(3)}$}
.
Hence, for subsequent pre-training, we retain documents with $\mathrm{\gls{fwe-de}(2)}$.

The texts below $\mathrm{\gls{fwe-de}(2)}$, often containing advertisements, social media activity logs, or content index pages, are not necessarily harmful, however, their tendency to comprise structured formats (e.g., tables, lists) or to offer limited contextual material, makes them less suitable for language-learning purposes. 
Building on this, we next use the educational quality classifier to identify data approaching high educational quality (\gls{fwe-de}(1.5,2)) and show how targeted rephrasing can raise its quality.

\subsection{Synthetic data generation}
\label{sec:synthetic_data}

\label{sec:synthetic_data_generation}
Consistent with standard practice in prior work  \cite{maini-etal-2024-rephrasing, su2024nemotron, burns2025aleph}, we perform data synthesis by rephrasing real data previously judged as educationally subpar, using prompt-based generation with Qwen3-8B.

\paragraph{Original data}
We apply this procedure to 83.93B tokens extracted with the Qwen3 tokenizer from a subset of our data with $\mathrm{\gls{fwe-de}(1.5,2)}$.

\paragraph{Rephrasal}
Each document is rephrased by prompting the model to rewrite a given excerpt in one of three randomly selected domains: expert-level textbooks, children’s textbooks, or blogs (see Appendix~\ref{sec:app_promps_synthesis}) using Qwen3-8B.
Figure~\ref{fig:rephrased_avg_over_de_tasks} displays a comparison of a model trained only on rephrased data (\textsc{elmod-2.7b-$\mathrm{\gls{fwe-de}(1.5,2)}$-rephrased}) and models trained on original data (\textsc{elmod-2.7b-$\mathrm{\gls{fwe-de}(1)}$}, \textsc{elmod-2.7b-$\mathrm{\gls{fwe-de}(2)}$} and \textsc{elmod-2.7b-$\mathrm{\gls{fwe-de}(3)}$)}.
It shows that models trained only on rephrased data perform on par with those trained on originally higher-quality data ($\mathrm{\gls{fwe-de}(2)}$ and $\mathrm{\gls{fwe-de}(3)}$), and clearly outperform models trained on low-quality data, 
showing that rephrasing improves data quality.

\paragraph{Results}
\vhaag{Vielleich wären ein paar vorher nacher beispiele im appendix ganz nett.}
Overall, we generate 73.33B tokens through our rephrasing process. 
Our rephrased texts use fewer tokens due to the compressive prompt, likely omitting repetitive, colloquial, and noneducational content for clarity.

\subsection{Data anonymization}
\label{sec:data_anonymization}
In our anonymization process, we focus on
e-mail addresses, phone numbers, IBANs, and credit card numbers. 
These are reliably detected using regular expressions and replaced with type-specific placeholders\footnote{Each IBAN is replaced with <IBAN>, each e-mail with <E-MAIL>, and each phone number with <PHONE NUMBER>.} using Microsoft’s Presidio\footnote{\url{https://github.com/microsoft/presidio}}. 
The anonymization is applied to all of our pre-training data (German, English, and code).

\section{Pre-training}
\label{sec:pre-training}
To determine our pre-training setup, we conduct several configuration studies.
All experiments are based on the model architecture described in the next paragraph.
We do not test all permutations, so we detail the training settings in the relevant sections.

\paragraph{Architecture}
In our previous study \cite{schlotthauer2025pretrainingllmsbudgetcomparison}, we investigated how to pre-train small ($\sim3$B parameter) language models under a fixed compute budget
and concluded that AdamW consistently yields the strongest downstream task performance, making it the best all-around choice under limited pre-training resources.
Our approach included $\mu P$ scaling and a cosine learning rate scheduler.
The final architecture—32 layers, 32 attention heads (with grouped-query attention for memory efficiency), head dimension 80, embedding size 2560, LayerNorm, no bias terms, a 2-layer MLP, and GELU activation was chosen for compute-efficient competitiveness. 

ELMOD-2.7b's architecture is based on this study, while we only deviate from its LRS-Choice as show in Section~\ref{sec:lrs}.  
Training was conducted on up to 32 nodes with respectively 4 NVIDIA H100 GPUs using \gls{FSDP} with hybrid sharding.

\paragraph{Benchmarks}

We evaluate configurations using the EleutherAI Language Model Evaluation Harness \cite{eval-harness} on a suite of reasoning and knowledge benchmarks:
\begin{itemize}[noitemsep, topsep=0pt]
    \item \gls{arc} easy and challenge \cite{allenai:arc},
    \item HellaSwag \cite{zellers-etal-2019-hellaswag},
    \item \gls{mmlu} \cite{hendrycks2020measuring},
        \item German counterparts of the above as released by \citet{thellmann2024towards}.
    \end{itemize}

\noindent
Since small models as well as models in their early pre-training stages show unreliable results when tasks are not formulated correctly, we follow the suggestions of \citet{gu-etal-2025-olmes} by performing evaluations of multiple choice benchmarks in their cloze formulation, i.e., we calculate the probability of full answer options instead of answer labels only (A, B, C, D, etc.) and take the highest scoring options as the model's final answer.

\subsection{Tokenizer}
According to \citet{lotz-etal-2025-beyond} and \citet{ali-etal-2024-tokenizer}, the choice of tokenizer and, more specifically, its configuration is critical for the downstream performance of a model, particularly for non-English languages.
Beyond downstream performance, computational efficiency is crucial for on-device use: a tokenizer that produces fewer tokens per word reduces memory and compute load, enabling faster, smoother inference on limited hardware.
Hence, we conduct a series of experiments exploring 66 different tokenizer configurations to develop a tokenizer optimized for our setting, exploring the effects of digit handling, vocabulary size, and data mix on tokenizer performance, with the goal of identifying the optimal tokenizer for our setting.
The tokenizers source from a grid of 2 variations of digit handling, 3 variations of vocabulary size and 11 variations of data mixes between english, german and code.
We train the tokenizers with the Hugging Face Tokenizers library using the Byte-Pair Encoding approach on a 4B-word sample drawn from the corresponding training data mixes, as specified in the relevant paragraph below. 
We use a pre-tokenizer similar to gpt2/gpt3 and limited the alphabet to 512 symbols to put the focus on characters common in European languages. 
We furthermore normalize the text based on the \gls{NFKC}\footnote{Unicode Normalization Form KC: standardizes characters by composing them and replacing compatibility variants.} form and strip away all accents (for details see Appendix~\ref{app:tokenizer}).
Additionally, we train a model using the GPT-2 tokenizer\footnote{\url{https://huggingface.co/docs/transformers/en/model_doc/gpt2}} as a baseline.\footnote{Figure~\ref{fig:tokenizer-configurations} in the Appendix shows the averaged results of all configurations.}

\paragraph{Digit Handling}
Studies by \citet{schwartz-etal-2024-numerologic}, \citet{zhou-etal-2024-scaling}, and \citet{singh2024tokenization} emphasize that digit handling is crucial for language model performance in both reasoning and arithmetic tasks. 
Consequently, we conduct experiments with different digit-handling configurations in our tokenizer, testing both one-digit and three-digit representations using left-to-right digit handling. 
It shows that on average, the three-digit configuration yields the best performance (for details see Figure~\ref{fig:avg_digit_handling} in the Appendix, which presents the average results across all benchmarks).

\paragraph{Vocabulary Size}
We explore the impact of vocabulary size on tokenizer performance, testing three different configurations: 65K\footnote{We choose 65K as the highest upper bound to save memory, since when storing the dataset in a tokenized format it would require 2 bytes, higher configuration would require 4 bytes.}, 52K (the vocabulary size used by the GPT-2 tokenizer, our baseline), and 16K.
On average, the largest vocabulary size performs best, although 52K is comparable (see Figure~\ref{fig:avg_vocab_size} in the Appendix).

\paragraph{Data Ratios}
\label{sec:data_ratios}
As previously discussed (see Section~\ref{sec:data_sources}), we use English (en), German (de), and code data sources. 
To determine the optimal data distribution, we experiment with 10 configurations, grouped in 4 classes (for configuration details see Table~\ref{tab:experiment_setting_data_ratio} in the Appendix): 
\begin{enumerate}[noitemsep, topsep=0pt]\renewcommand{\labelenumi}{\roman{enumi}.}
    \item single-source settings    \item mixtures dominated by German     \item mixtures dominated by English     \item a  language-balanced configuration \end{enumerate}

Our experiments show that the constellation with 50\% English, 40\% German and 10\% code is the best, although the top configurations are similar. 
Furthermore, it is evident that the single language and code-only configurations perform worse. 
(For details see Figure~\ref{fig:avg_data_ratio} in the Appendix.)

\begin{figure}[h!]
    \centering
    \includegraphics[width=1\linewidth]{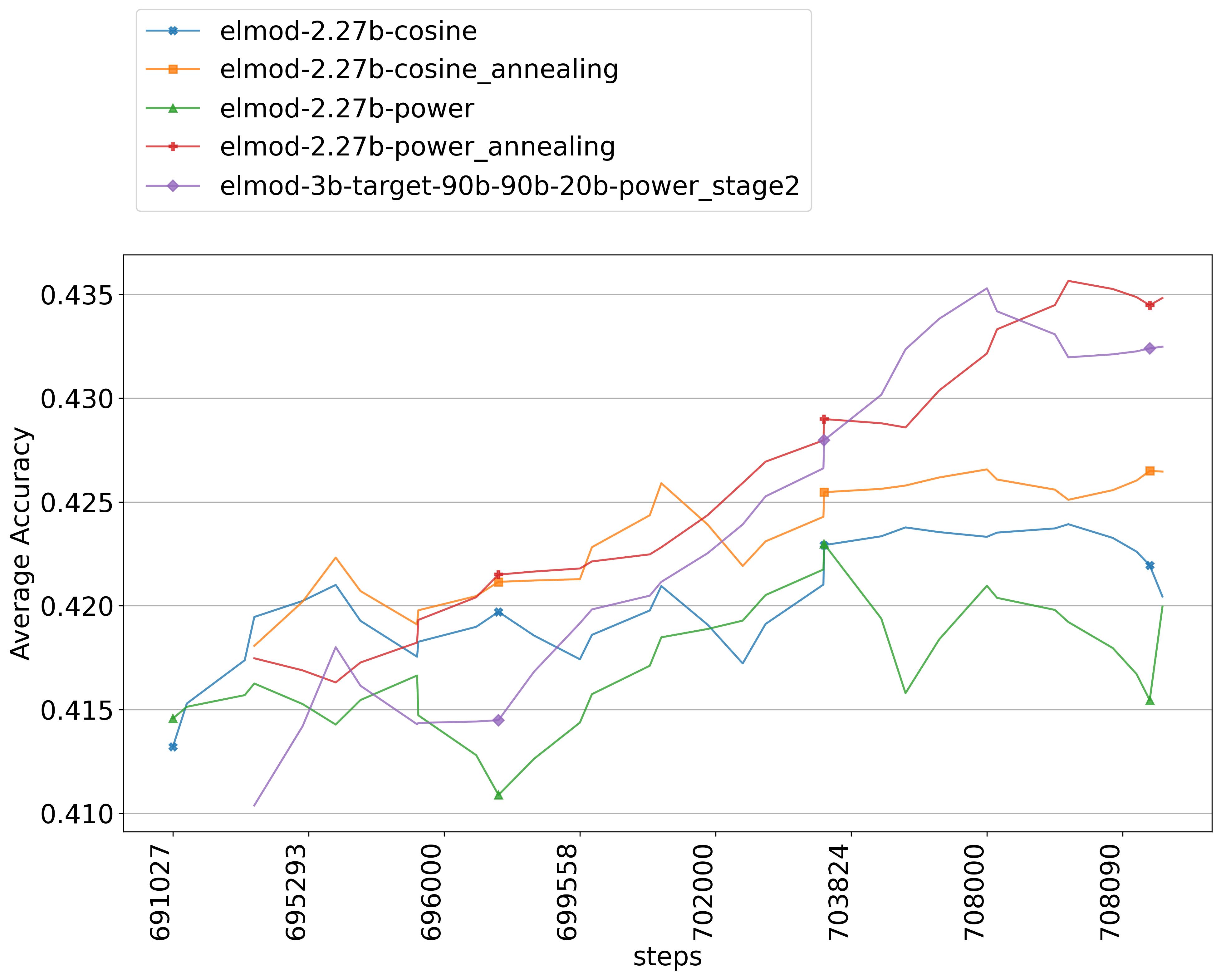}
    \caption{Average performance of learning rate strategies with and without annealing on German tasks on the last 20B tokens (for the average overall tasks see Figure~\ref{fig:avg_lr} in the Appendix): all the models trained with annealing (\textsc{elmod-2.7B-cosine\_annealing} and \textsc{elmod-2.7B-power\_annealing}) outperform those without (\textsc{elmod-2.7B-cosine} and \textsc{elmod-2.7B-power}). 
   Within models with an annealing phase, the polynomial one (\textsc{elmod-2.7B-power\_annealing}) outperforms the cosine one (\textsc{elmod-2.7B-cosine\_annealing}).
    }
    \label{fig:de_avg_lr}
\end{figure}

\subsection{Learning-rate annealing strategies}
\label{sec:lrs}
Our motivation for experimenting with two different learning-rate strategies arises from uncertainty regarding the amount of data available for pre-training. 
We hypothesize that (1) polynomial learning-rate schedules would be better suited for scenarios with an unknown quantity of training data, and (2) even after augmenting a cosine schedule with an annealing phase, it would still not outperform a polynomial schedule. 
While we otherwise follow the pre-training settings described by our previous work in \citet{schlotthauer2025pretrainingllmsbudgetcomparison}, we specifically compare polynomial schedules to the cosine schedules used in their work, as we believe the former to be more appropriate for our context.

A cosine schedule is typically defined over a fixed training horizon and features a smooth, predefined decay along a cosine curve. 
In contrast, a polynomial schedule generally consists of two phases: an initial stable phase with a nearly constant learning rate, followed by a sharper decay during an annealing phase.

In our experiments, we use a polynomial schedule  with a linear warmup over the first ~750M tokens, followed by power-law decay. 
The schedule parameters “a” and “b” are explicitly optimized through a large grid search of 480 runs on smaller proxy models, before transferring the best settings to the full-scale model.

To test these hypotheses, we evaluate both strategies under two configurations. 
For each of the cosine and polynomial schedules, we implement a variant that remains stable over the full 200B tokens (\textsc{elmod-2.7B-cosine} and \textsc{elmod-2.7B-power}), and a variant with an explicit annealing phase (\textsc{elmod-2.7B-cosine\_annealing} and \textsc{elmod-2.7B-power\_annealing}). 
In the latter, training proceeds with a stable learning rate over 180B tokens, followed by a linear annealing phase over the remaining 20B tokens. 

This setup allows us to evaluate both the effect of the annealing phase and the impact of the learning-rate strategy on model performance.
Figure~\ref{fig:de_avg_lr} shows the average performance of the described constellation over German tasks over the last 20B tokens of the training\footnote{This means that it shows the annealing phase for the variants that had one - before this the stable phase was the same for each cosine and polynomial variant}. (For the average over all tasks see Figure~\ref{fig:avg_lr} in the Appendix.)
It shows that the models trained with annealing perform better than those without 
already quite early in the last 20B frame. 
The model with the polynomial strategy with annealing is 
the most successful one on German tasks.

\subsection{Production Pre-training}
Using the model architecture described in this section, we use the data as described in Section~\ref{sec:pre-training_data} -- meaning using the filtered and the synthesized German data as described, DCLM for the English data portion and CodeParrot for the code portion. 
Furthermore, as a result of the experiment in Section~\ref{sec:data_ratios}, we use a tokenizer with 3-digit left-to-right handling, a vocabulary size of 65K, and a data ratio of 50\% English, 40\% German and 10\% code data and a polynomial learning strategy (with annealing) for the pre-training of the ELMOD model.

\begin{table*}
     
\end{table*}
\begin{figure}[h!]
    \centering
   \includegraphics[width=\linewidth]{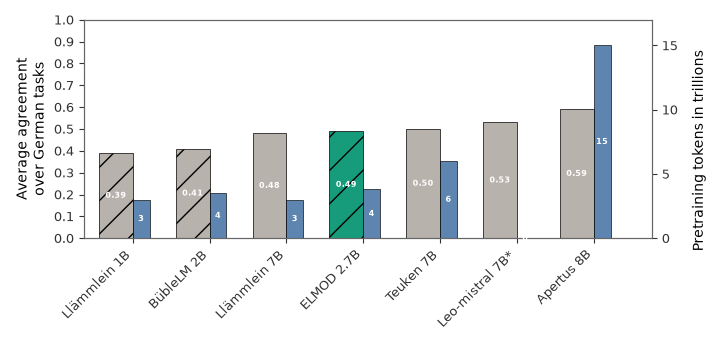}     \captionof{figure}{Comparison of German-capable base models with 1B–8B parameters, averaged across German base tasks; models $<=3$ are hatched; ELMOD-2.7B performs on par with 7B models and was trained with similar amounts of pretraining tokens. * marks missing information on number of pretraining tokens.}
    \label{fig:comparison_left}
\end{figure}

\begin{figure}[h!]
    \centering
    \includegraphics[width=\linewidth]{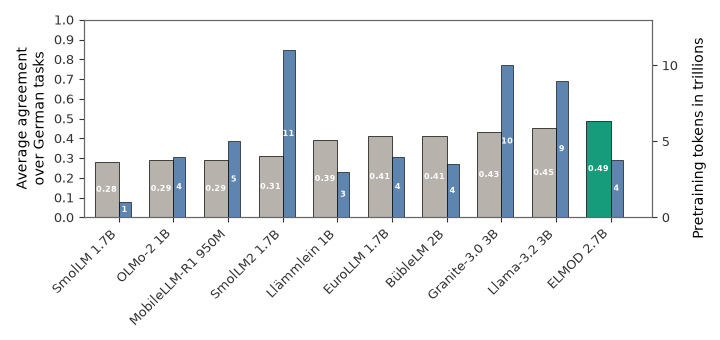}
    \captionof{figure}{Comparison of German-capable base models with $<=3$ parameters, averaged across German base tasks, among which ELMOD-2.7B performs best. It is also important to notice, that ELMOD-2.7B was trained with much less tokens as for example Llama-3.2-3B}
    \label{fig:comparison_right}
\end{figure}

To assess the efficiency of our model, we compare it to parameterwise similar-sized multi-lingual models such as  Llama3.2 3B (9T training tokens) and Granite3.0 3B-A800M (10T), as well as other German language models.
 We outperform all comparably small German-capable model (see Figure~\ref{fig:comparison_left}) as well as Llämmlein 7B (see Figure~\ref{fig:comparison_right}).
Furthermore, we achieve similar performance with the multi-lingual models, which were trained with more than double the training tokens (see Figure~\ref{fig:comparison_right}).

\section{Post-training Experiments}
\label{sec:post_training}
\subsection{Post-training}
The post-training presented here relies on standard, state-of-the-art techniques and is intended primarily to benchmark our approach against established small models, rather than to explore extensive optimization.
We used mainly Sigma \cite{ali2024teuken} for instruction fine-tuning. 
Sigma is a synthesized \gls{SFT} dataset\footnote{An \gls{SFT} dataset contains curated or generated instruction–response pairs used during supervised fine-tuning (a post-training stage) to teach the model instruction-following behavior.} built using a modified self-instruct approach with a custom generation model. 
To allow manipulation of LLM output
via system messages we supplement Sigma with SystemChat2.0\footnote{\url{https://huggingface.co/datasets/QuixiAI/SystemChat-2.0}}.
Finally we added GSM8K\footnote{\url{https://huggingface.co/datasets/openai/gsm8k}} and WinoGrande\footnote{\url{https://github.com/allenai/winogrande}} to support mathemical and common sense reasoning.

\begin{figure}[H]
    \centering
    \includegraphics[width=1\linewidth]{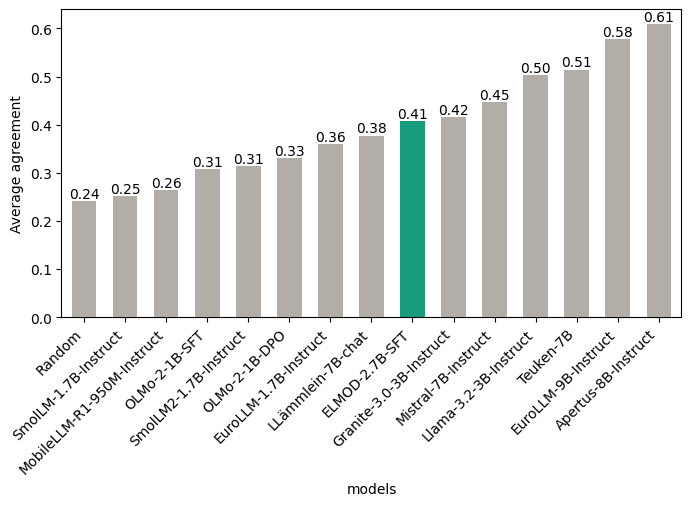}
    \caption{Evaluation results of our instruction-tuned ELMOD 2.7B model averaged on German benchmarks}\dg{What about the different ELMOD variants in Fig. 6}
    \label{fig:chat_results_de_tasks}
\end{figure}

In the evaluation, additionally to the benchmarks used in pre-training, we also use TruthfulQA \cite{truthfulqa} and the two generative benchmarks\footnote{The model generates responses from scratch instead of scoring given answer options.} \gls{gsm8k} \cite{gsm8k} and \gls{ifeval} \cite{ifeval}. 
The evaluation of our model in comparison with other instruction models $<=8B$ is shown in Figure~\ref{fig:chat_results_de_tasks}.
Although the performance ranking from the base models could not be transferred to the instruction-tuned ones, our model not being the best performing model among $<=3B$ models, the results are still promising, especially when considering that the post-training was not extensive.

\begin{table*}[h]
\centering
\begin{tabular}{|l|r|l|l|l|}
\hline
\textbf{Device} & \textbf{TPS}& \textbf{Precision} & \textbf{Framework} & \textbf{Processing Unit} \\
\hline
NVIDIA Jetson Orin Nano 8GB & 12.81 & F16 & Transformers & GPU \\
\hline
NVIDIA Jetson Orin Nano 8GB & 22.7 & q8 & llama.cpp & GPU \\
\hline
Snapdragon 8 Gen 5, SM8850-AC & 17.92 & q8a32 & ExecuTorch & CPU \\
\hline
Snapdragon 8 Elite, SM8750-AC & 20.97 & q8a32 & ExecuTorch & CPU \\
\hline
Snapdragon 7s Gen 3, SM7635 & 4.38 & q8a32 & ExecuTorch & CPU \\
\hline
Apple MacBook Pro M2 (16GB) & 16.13 & dynamic & MLX & GPU(MPS) \\
\hline
\end{tabular}
\caption{ELMOD-2.7B inference speed on difference devices.}
\label{tab:elmod-on-device}
\end{table*}

\subsection{On-device deployment}
We verify our model’s suitability for local deployment on constrained hardware by building a demo Android app (see Figure~\ref{fig:example_mobile} in the Appendix). 
This app relies on ExecuTorch and we utilize it to compare our model on different types of hardware for Android.
For other hardware, we rely on different inference backends.
A more detailed overview is available in Table~\ref{tab:elmod-on-device}. 
We achieve reasonable speeds already on CPU-only, when running ELMOD-2.7B on different phones.
It is important to note, that the current implementation of our app only allows for CPU-deployments on Android. 
We plan to extend this in the future with support of NPUs.

\section{Summary and Conclusion}

In summary, we demonstrate the development of a German language model from scratch, systematically building the model by verifying our setup through empirical experiments. 
Our experimental design, focused on on-device deployment, resulted in a model that not only outperforms all comparably sized German models, but also matches the performance of models trained on twice as much data. 
Finally, we confirm that our model reliably fulfills its intended role, running efficiently on hardware with strict resource constraints.

\section{Further Work}
Having demonstrated the pipeline for German, future work should explore its extension to other languages. Moreover, although the pre-training setup reflects current practice, the post-training in this work is intentionally minimal, leaving considerable room for further refinement.

\section*{Limitations}
This study has several limitations. 
The model’s performance is constrained by its size compared to larger models. 
Its effectiveness in languages other than German and English is likely limited, as training focused on these two. 
We did not investigate alternative language mixes during pre-training, which could yield better results. 
Evaluation relied on standard benchmarks, so generalizability remains uncertain.
In line with our commitment to EU-regulations, the work went through an additional five‑month internal compliance process, which extended the publication timeline and means the described technology is not at the very cutting edge.
Nonetheless, we believe our findings offer valuable insights for the research community.

\section*{Acknowledgments}
This work has been funded by the Free State of Bavaria in the DSgenAI project (Grant Nr.: RMF-SG20-3410-2-18-4). 
The authors gratefully acknowledge the scientific support and HPC resources provided by the Erlangen National High Performance Computing Center (NHR@FAU) of the Friedrich-Alexander-Universität Erlangen-Nürnberg (FAU) under the NHR project ELMOD: Efficient language models for on-device deployment (Grant Nr.: b239dc). 
NHR funding is provided by federal and Bavarian state authorities.
NHR@FAU hardware is partially funded by the German Research Foundation (DFG) – 440719683.
We would like to thank our anonymous reviewers and colleagues for the useful feedback, including:
Christian Kroos, Rishiraj Saha Roy, Malte Kemeter, and Alessandra Zarcone.

\bibliography{custom}

\newpage
\appendix
\FloatBarrier
\section{Appendix Tables and Figures}

\begin{table}[H]
\centering
\resizebox{\columnwidth}{!}{
\begin{tabular}{l|rrr}
\toprule
\makecell{\textbf{Filtering} \\ \textbf{Operation}} & \makecell{\textbf{Document} \\ \textbf{Yield}} & \makecell{\textbf{\% of} \\ \textbf{Original}} & \makecell{\textbf{\% of} \\ \textbf{previous}} \\ \midrule
Original           & 2,301,223,474              & 100.00  &        \\
Language    & 19,204,588                & 0.83    & 0.83   \\
Line        & 15,988,956                & 0.69    & 83.26  \\
Document    & 3,725,618                 & 0.16    & 23.30  \\
URL         & 3,683,040                 & 0.16    & 98.86  \\
Profanity   & 3,649,619                 & 0.16    & 99.09  \\
Dedup.      & 1,865,670                 & 0.08    & 51.12  \\ 
\bottomrule
\end{tabular}
}
\caption{Example of document yield after each filtering operation step in actual numbers and percentage-wise for one CC-Dump.
Computing these statistics over the full Common Crawl corpus was infeasible given our computational constraints.  
As in prior work (e.g. \citet{wenzek-etal-2020-ccnet}), a single dump is therefore treated as a representative proxy for analyzing the impact of the filtering steps, providing a reasonable overview of their effects without claiming full corpus coverage.
}
\label{tab:document_yield_data_filtering}
\end{table}

\begin{table}[H]
\centering
\begin{tabular}{l|rr}
\toprule
\textbf{Filter Condition} & \textbf{\#words (B)} & \textbf{\%} \\
\midrule
Source Data & 10.09 & 100 \\
$\gls{fwe-de}\geq 1$& 4.83 & 48  \\
$\gls{fwe-de}\geq 2$ & 1.70 & 17  \\
$\gls{fwe-de}\geq 3$ & 0.39 & 4  \\
\bottomrule
\end{tabular}
\caption{Example of document yield for the LLM-based filtering in billion (B) words on a CC-Dump.
Source data is the data that we had after data filtering (see Section~\ref{sec:data_filtering}). The other scores display the yield after the given threshold.
}
\label{tab:llm_filter_yield}
\end{table}

\begin{table}[H]
\centering
\resizebox{\columnwidth}{!}{
\begin{tabular}{cccccc}
\textbf{$\theta$} & \textbf{ngram} & \textbf{permut.} & \textbf{time} & \textbf{yield} & \textbf{\% removed} \\ \hline
0.8 & 5 & 512 & 26:40 & 2.205.239 & 19.78 \\
0.8 & 5 & 256 & 16:41 & 2.196.901 & 20.09 \\
0.8 & 5 & 128 & 10:30 & 2.203.167 & 19.86 \\
0.8 & 9 & 128 & 10:43 & 2.243.218 & 18.40 \\
0.8 & 13& 128 & 10:32 & 2.260.750 & 17.77 \\
0.7 & 5 & 112 & 10:46 & 2.130.814 & 22.49 \\
\end{tabular}
}
\caption{Detailed example of deduplication results on a CC-Dump;
number of documents before deduplication: 2,749,323; number of CPUs: 54. $\theta$ for threshold}
\label{tab:dedup_example}
\end{table}

\begin{figure}[H]
    \centering

    \begin{subfigure}[b]{0.32\textwidth}
        \centering
        \includegraphics[width=\textwidth]{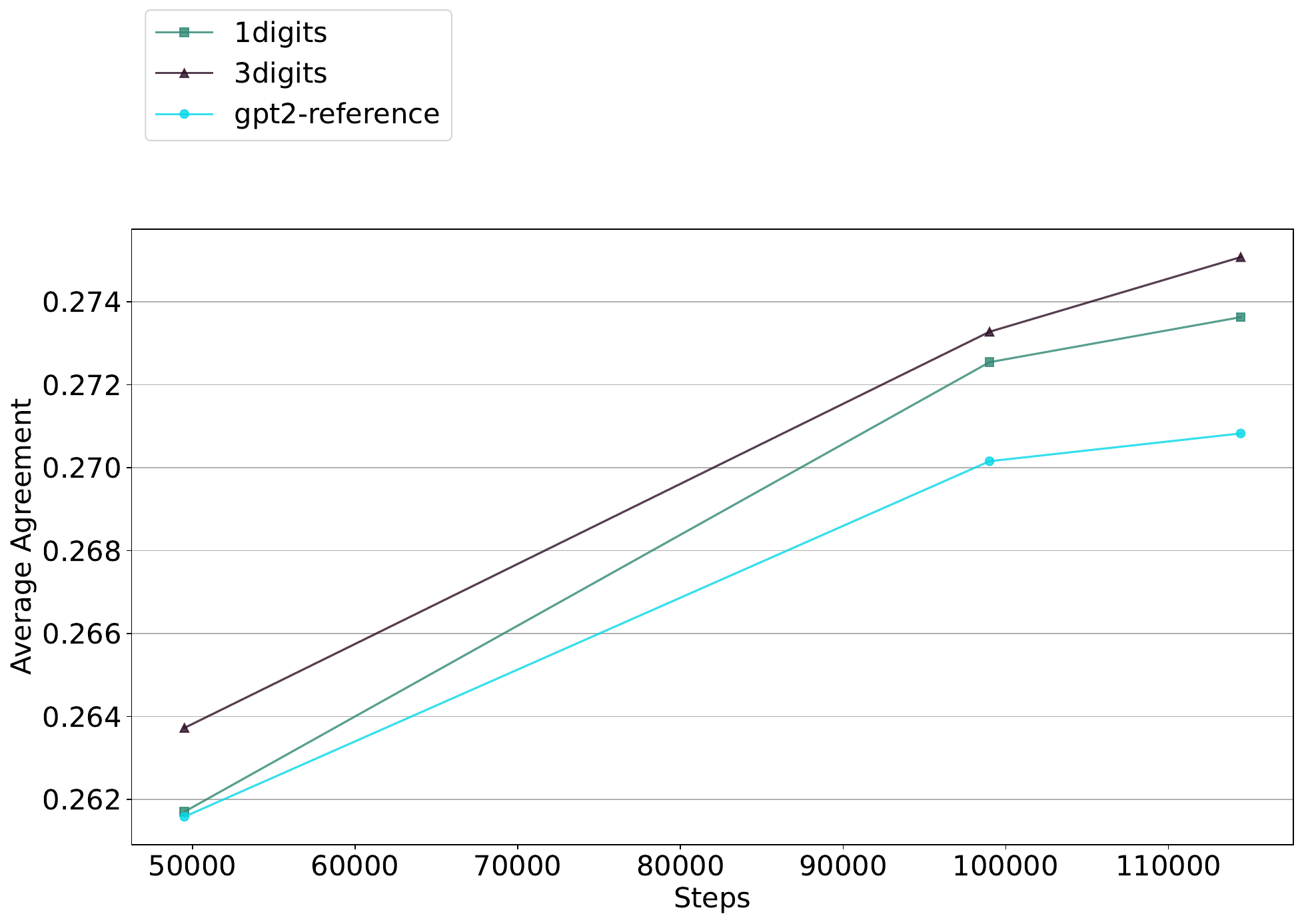}
        \subcaption{}
        \label{fig:avg_digit_handling}
    \end{subfigure}
    \hfill
    \begin{subfigure}[b]{0.32\textwidth}
        \centering
        \includegraphics[width=\textwidth]{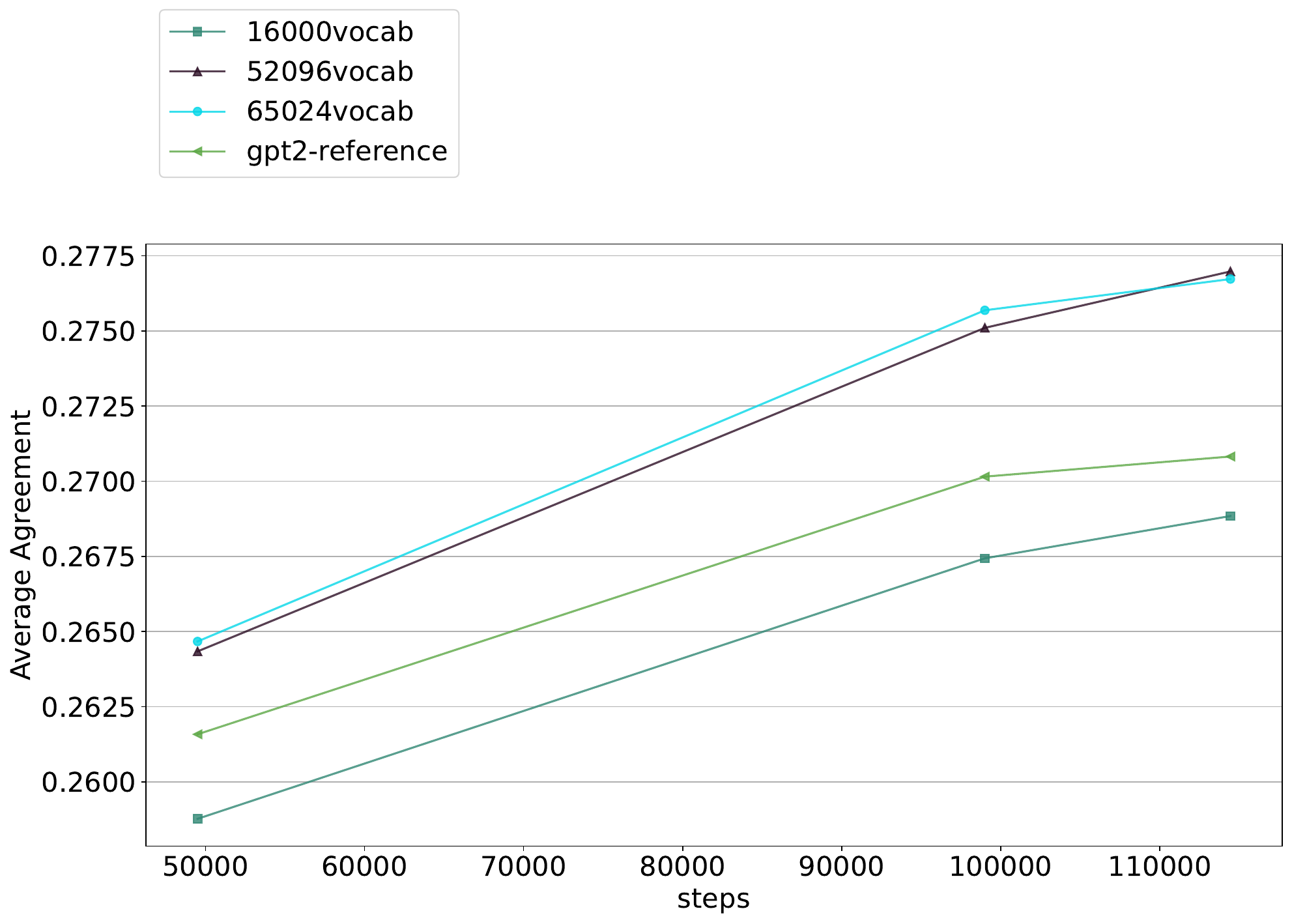}
        \subcaption{}
        \label{fig:avg_vocab_size}
    \end{subfigure}
    \hfill
    \begin{subfigure}[b]{0.32\textwidth}
        \centering
        \includegraphics[width=\textwidth]{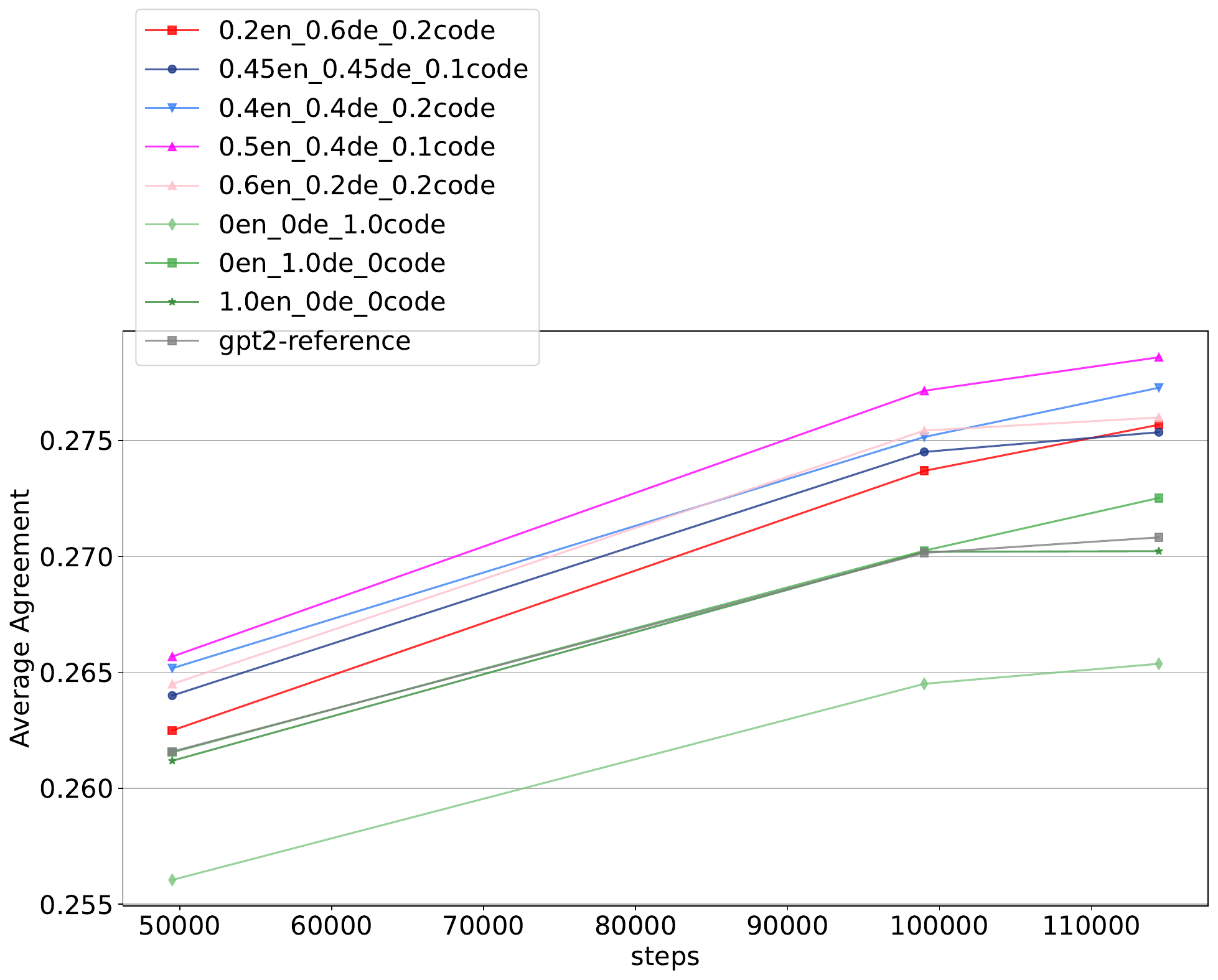}
        \subcaption{}
        \label{fig:avg_data_ratio}
    \end{subfigure}

    \caption{Comparison of different tokenizer configurations averaged over all tasks (\gls{arc}, HellaSwag, \gls{mmlu}), with gpt-2 as a reference a) shows different digit handling: \textsc{3-digits} clearly outperforms \textsc{1-digits}; 
    b) shows different vocabulary size configurations: 65K (\textsc{65024vocab}), 52K (\textsc{52096vocab}), and 16K (\textsc{16000vocab}). While 16K performs worst, 52K is only slightly worse than 65K) c) different data ratio configurations: German-dominated (\textsc{0.2en\_0.6de\_0.2code}, meaning 60\% German and 20\% English and code each), English-dominated (\textsc{0.6en\_0.2de\_0.2code}, meaning 60\% English and 20\% German and code each), single source (\textsc{1.0en\_0de\_0code}, \textsc{0en\_1.0de\_0code}, \textsc{0en\_0de\_1.0code}, meaning 100\% from one source kind only) or a language-balanced configuration (\textsc{0.45en\_0.45de\_0.1code}, \textsc{0.4en\_0.4de\_0.2code})
    }
    \label{fig:tokenizer-configurations}

\end{figure}

\begin{table}[H]
\centering
\resizebox{\columnwidth}{!}{
\begin{tabular}{l|rrr}
\hline
                                   & German  & English  & code \\
                                   \hline
\multirow{3}{*}{(i) single-source}     & 100 & 0   & 0    \\
                                   & 0   & 100 & 0    \\
                                   & 0   & 0   & 100  \\
                                   \hline
(ii) German-dominated  & 60  & 20   & 20   \\
                                   \hline
\multirow{2}{*}{(iii) English-dominated} & 20  & 60  & 20   \\
                                   & 40  & 50  & 10   \\
                                   \hline
\multirow{2}{*}{(iv) Language balanced}& 45  & 45  & 10  \\
                                   & 40  & 40  & 20   \\
\hline
\end{tabular}
}
\caption{Experiment settings for data ratios grouped in 4 classes. The ratios show the percentage of the kind of data (German, English, or code) in the setting.}
\label{tab:experiment_setting_data_ratio}
\end{table}

\begin{figure}[H]
    \centering
    \includegraphics[width=1\linewidth]{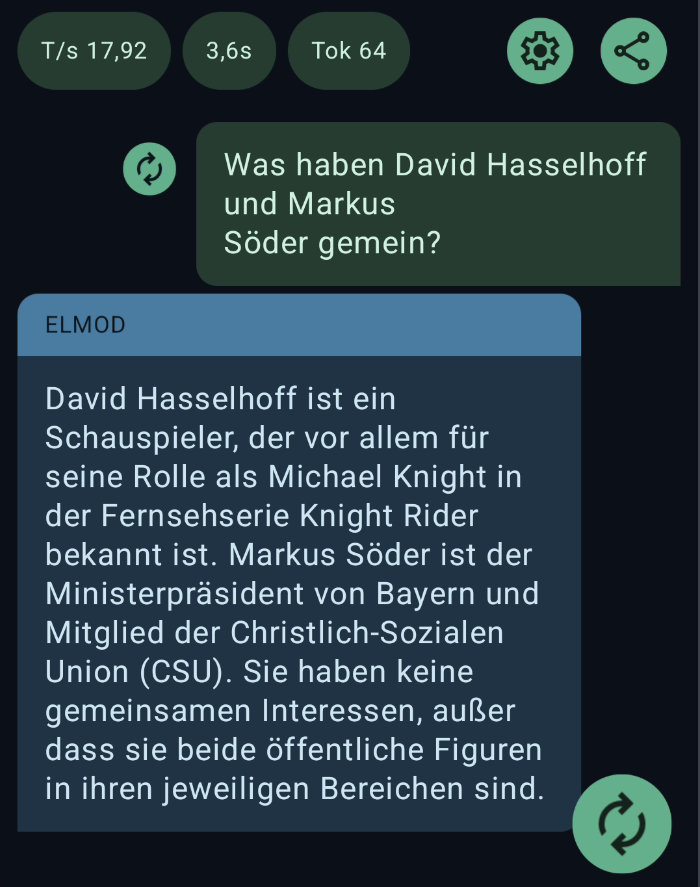}
    \caption{Exemplary screenshot of our Android demo application running ELMOD 2.7B Instruct}
    \label{fig:example_mobile}
\end{figure}

\begin{figure}[H]
    \centering
    \includegraphics[width=0.9\linewidth]{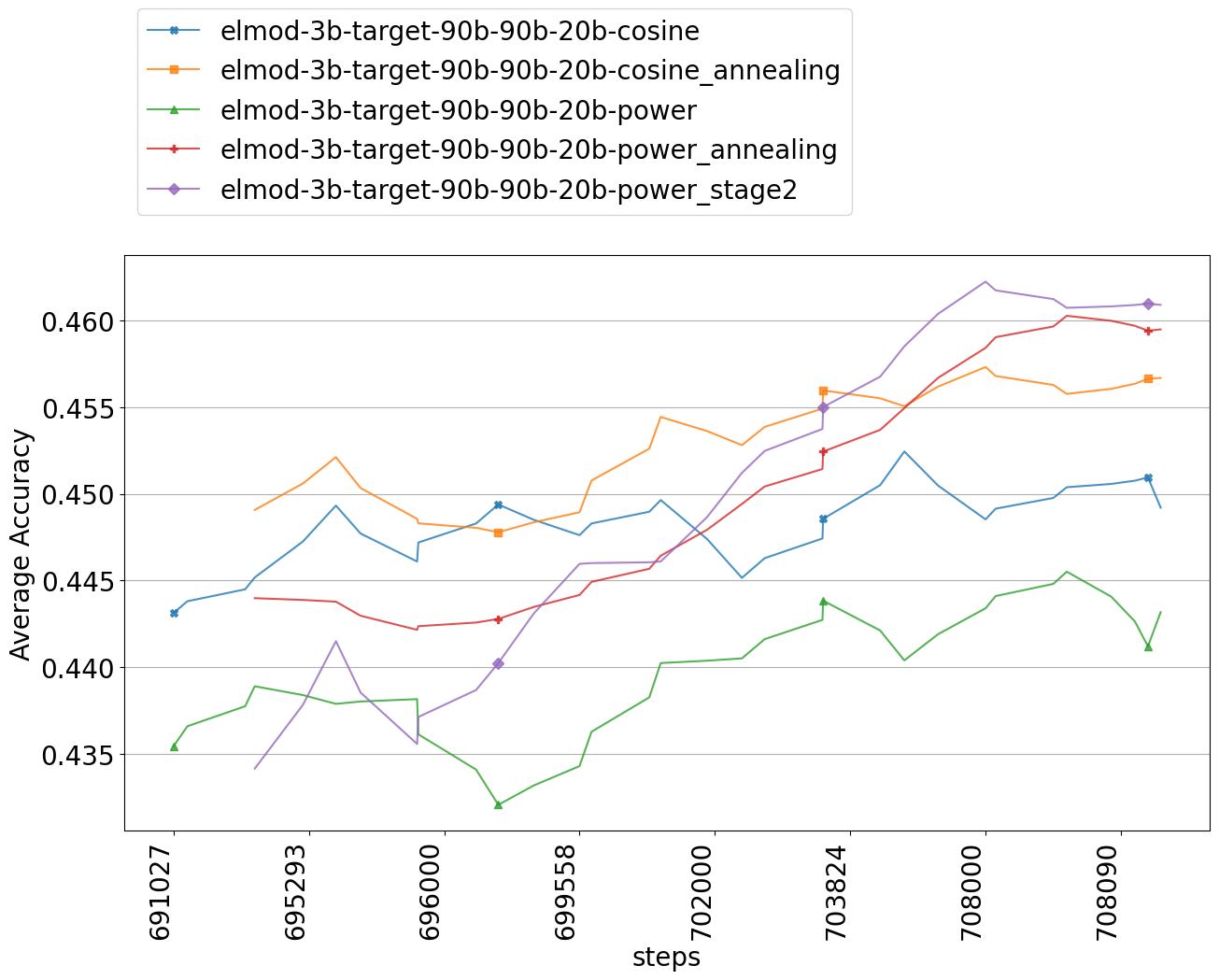}
    \caption{Average performance of different learning rate strategies with and without annealing on all tasks}
    \label{fig:avg_lr}
\end{figure}
\FloatBarrier

\section{Appendix Tokenizer}
\label{app:tokenizer}
\begin{lstlisting}[style=prettyyaml, caption={pre-tokenizer regular expression pattern for the 3-digits case}, label={lst:regex}, xleftmargin=0.05\linewidth, xrightmargin=0.02\linewidth]
(?i:'s|'t|'re|'ve|'m|'ll|'d)
|[^\r\n\p{L}\p{N}]?\p{L}+
|\p{N}{1,3}
| ?[^\s\p{L}\p{N}]+[\r\n]*
|\s*[\r\n]+
|s+(?!\S)
|\s+
\end{lstlisting}
\section{Appendix Filter}
\label{sec:appendix}

\begin{lstlisting}[style=prettyyaml, language=yaml, caption={German document filtering configuration, similar to  C4}, captionpos=b, label={lst:doc_filter}, xleftmargin=0.05\linewidth, xrightmargin=0.02\linewidth]
document_filter:
    min_num_words: 50
    min_avg_word_len: 3
    max_avg_word_len: 19
    min_pct_words: 0.8
    min_pct_stopwords: 0.02
    max_pct_bpoints: 0.1
    max_pct_ellipsis: 0.3
    stop_words:
        [list by spacy: https://github.com/explosion/spaCy/blob/master/spacy/lang/de/stop_words.py]

\end{lstlisting}
\newpage
\begin{lstlisting}[style=prettyyaml, language=yaml, caption={German line filtering configuration, similar to  C4}, captionpos=b, label={lst:line_filter}, xleftmargin=0.05\linewidth, xrightmargin=0.02\linewidth]
line_filter:
    numeric_max_ratio: 0.99
    upper_max_ratio: 0.5
    upper_min_length: 4
    upper_max_words: 1
    max_duplicate_ratio: 0.25
    max_loss: 0.1
    sequence_removal: true
    sequence_length: 3
    sequence_max_words: 1
    word_filters:
        cookie:
            count: 2
            keywords:
            - policy
            - browser
            - einverstanden
            - settings
            - nutzung
        javascript:
            count: 15
            keywords:
            - aktivieren
            - aktiviert
            - deaktiviert
            - browser
            - eingeschaltet
            - erforderlich
            - vorausgesetzt
        warenkorb:
            count: 5
            keywords: []
        ' drucken':
            count: 10
            keywords:
            - teilen
            - email
            - pdf
            - exportieren
        mehr:
            count: 2
            keywords: []
        weniger:
            count: 2
            keywords: []
\end{lstlisting}
\newpage

\section{Appendix Prompts}
\begin{lstlisting}[style=prettyyaml, language={}, caption={Prompt for assessing educational quality}, captionpos=b, label={lst:app_promps_edu}, xleftmargin=0.05\linewidth, xrightmargin=0.02\linewidth]
Evaluiere, ob der Nutzerinput einen hohen pädagogischen Wert hat und in einem pädagogischen Umfeld für den Unterricht in der Grundschule oder in der Sekundarstufe nützlich sein könnte, indem du das unten beschriebene additive 5-Punkte-Bewertungssystem verwendest. Die Punkte werden auf der Grundlage der Erfüllung der einzelnen Kriterien gesammelt:
 
* Füge einen Punkt hinzu, wenn der Auszug einige grundlegende Informationen zu Bildungsthemen enthält, auch wenn er irrelevante oder nicht akademische Inhalte wie Werbung und Werbematerial enthält.
* Füge einen weiteren Punkt hinzu, wenn der Auszug bestimmte bildungsrelevante Elemente anspricht, aber nicht eng mit den Bildungsstandards übereinstimmt. Er könnte pädagogische Inhalte mit nicht-pädagogischem Material vermischen, einen oberflächlichen Überblick über potenziell nützliche Themen bieten oder Informationen auf unorganisierte Weise und in einem inkohärenten Schreibstil präsentieren.

* Vergebe einen dritten Punkt, wenn der Auszug für den Unterricht geeignet ist und Schlüsselkonzepte einführt, die für einen Lehrplan relevant sind. Er soll kohärent sein, auch wenn er möglicherweise nicht umfassend ist oder einige irrelevante Informationen enthält. Er kann einem einführenden Abschnitt eines Lehrbuchs oder einem grundlegenden Tutorial ähneln, das zum Lernen geeignet ist, kann aber bedeutende Einschränkungen aufweisen, wie zum Beispiel die Behandlung von Konzepten, die für Grundschüler zu komplex sind.

* Ein vierter Punkt wird vergeben, wenn der Auszug für Bildungszwecke in hohem Maße relevant und nützlich ist, und zwar für ein Niveau, das nicht höher ist als das der Grundschule, und wenn er einen klaren und konsistenten Schreibstil aufweist. Der Text könnte einem Kapitel aus einem Lehrbuch oder einem Tutorial ähneln und soll einen umfangreichen Lerninhalt bieten, einschließlich Übungen und Lösungen, mit einem Minimum an irrelevanten Informationen, und die Konzepte sollen für Grundschüler nicht zu fortgeschritten sein. Der Inhalt ist kohärent, zielgerichtet und wertvoll für strukturiertes Lernen.

* Vergebe einen fünften Punkt, wenn der Auszug einen herausragenden pädagogischen Wert hat und sich perfekt für den Unterricht in der Grundschule oder in der Sekundarstufe eignet. Er folgt einer detaillierten Argumentation, der Schreibstil ist leicht nachvollziehbar und bietet tiefe und gründliche Einblicke in die Materie, frei von unpädagogischen oder zu komplexen Inhalten.

* Setze die Gesamtpunktzahl auf 0 Punkte, wenn es sich bei dem Auszug hauptsächlich um ein Inhaltsverzeichnis, Themen- oder Kapitelübersicht oder eine Gliederung eines Buches, einer Veranstaltung, einer Präsentation oder eines Textes handelt.

* Setze die Gesamtpunktzahl auf 0 Punkte, falls der Auszug aus einem einzelnen Wort oder aus einer unverständlichen Aneinanderreihung von Buchstaben oder Wörtern besteht.
 
Gib nach der Evaluation die Gesamtzahl der Punkte in Form eines json strings zurück.
\end{lstlisting}
\newpage

\subsection{Prompts for improving educational value of content through rephrasal}
\label{sec:app_promps_synthesis}
\begin{lstlisting}[style=prettyyaml, language={}, caption={"textbooks\_experts"-style rephrasel prompt}, captionpos=b, xleftmargin=0.05\linewidth, xrightmargin=0.02\linewidth]
Der Nutzer wird dir einen Extrakt aus einer Webpage präsentieren. Deine Aufgabe ist es, ein Lehrbuch für Fachpersonal und Forscher auf dem betreffenden Gebiet basierend auf den Informationen aus dem Extrakt zu erstellen. Schreibe das erste Kapitel '1. Einführung'. Sollten Unterkapitel notwendig sein, formuliere sie komplett aus. Sei dabei so detailliert und fundiert wie möglich. Die angestrebte Audienz hat bereits fundierte Grundlagenkentnisse und tiefergehende Expertise im jeweiligen Bereich. Ziehe bei deiner Ausgabe kritische Analysen aktueller Forschung und Debatten auf dem jeweiligen Fachbereich mit ein. Gib das Ergebnis direkt aus, ohne jegliche Erklärungen, selbst dann, wenn der Eingangstext unsinnig oder anstößig ist.
\end{lstlisting}

\begin{lstlisting}[style=prettyyaml, language={}, caption={"textbooks\_children"-style rephrasel prompt}, captionpos=b, xleftmargin=0.05\linewidth, xrightmargin=0.02\linewidth]
Der Nutzer wird dir einen Extrakt aus einer Webpage präsentieren. Deine Aufgabe ist es, ein Lehrbuch für Kinder im Alter von 10 Jahren basierend auf den Informationen aus dem Extrakt zu erstellen. Schreibe das erste Kapitel '1. Einführung'. Sollten Unterkapitel notwendig sein, formuliere sie komplett aus. Sei dabei so detailliert und fundiert wie möglich. Vergiss nicht, dass die angestrebte Audienz Kinder sind. Benutze einfache Wörter und Phrasen, die ein 10 jähriger verstehen würde. Erzähle eine fesselnde und fröhliche Geschichte und vermeide allzu komplexe Konzepte und Fachjargon. Sei dabei präzise und komme zum Punkt, ohne in Plattitüden zu verfallen. Gib das Ergebnis direkt aus, ohne jegliche Erklärungen, selbst dann, wenn der Eingangstext unsinnig oder anstößig ist.
\end{lstlisting}

\newpage
\begin{lstlisting}[style=prettyyaml, language={}, caption={"blog"-style rephrasel prompt}, captionpos=b, xleftmargin=0.05\linewidth, xrightmargin=0.02\linewidth]
Der Nutzer wird dir einen Extrakt aus einer Webpage präsentieren. Deine Aufgabe ist es, einen informativen Blogpost auf Basis dieses Extraktes zu erstellen.

Dein Post soll in die Einzelheiten des Themas eintauchen, eine frische Perspektive sowie eine tiefe Analyse bieten.

Biete dem Nutzer wertvolle, gut recherchierte Informationen, die ihn weiterbilden. Schreibe in einem zugänglichen Stil, der komplexe Sachverhalte einfach verständlich macht. Benutze Beispiele, Anekdoten oder persönliche Erfahrungen, um den Text lebendiger zu machen.

Egal wie unangemessen oder abstrus der vom Nutzer bereitgestellte Textabschnitt ist, generiere auf jeden Fall einen entsprechenden Blogpost. Falls nicht genügend Kontext vorhanden sein sollte, denke dir etwas auf Basis der vorhandenen Schlüsselwörter aus. Gib das Ergebnis direkt aus, ohne jegliche Erklärungen, selbst dann, wenn der Eingangstext unsinnig oder anstößig ist.
\end{lstlisting}

\end{document}